\documentclass[letterpaper,10pt,conference]{IEEEtran}

\IEEEoverridecommandlockouts

\usepackage[utf8]{inputenc}
\usepackage[T1]{fontenc}
\usepackage{bm}
\usepackage{amsmath}
\usepackage{amssymb}
\usepackage{graphicx}
\usepackage{cite}            % for compressed, sorted citations like [1]–[3]
\usepackage{placeins}
\usepackage{dblfloatfix}
\usepackage[pdfencoding=auto,psdextra]{hyperref} % UTF-8 안전

\hypersetup{
  pdftitle   = {HyperspectralMAE: The Hyperspectral Imagery Foundation Model using Fourier-Encoded Dual-Branch Masked Autoencoder},
  pdfauthor  = {Wooyoung Jeong and Hyun Jae Park and Seonghun Jeong and Jong Wook Jang and Tae Hoon Lim and Dae Seoung Kim},
  pdfsubject = {Self-supervised foundation model for hyperspectral imagery using dual-branch MAE},
  pdfkeywords= {foundation model, hyperspectral, masked autoencoder, transformer, remote sensing},
  pdfcreator = {LaTeX with hyperref},
  pdfproducer= {TeX Live 2024},
  breaklinks = true
}

\title{\LARGE \bf
HyperspectralMAE: The Hyperspectral Imagery Classification Model using Fourier-Encoded Dual-Branch Masked Autoencoder
}

\author{%
  \IEEEauthorblockN{%
    Wooyoung Jeong,
    Hyun Jae Park,
    Seonghun Jeong,
    Jong Wook Jang,
    Tae Hoon Lim,
    Dae Seoung Kim\textsuperscript{*}}
  \IEEEauthorblockA{%
    {AI Research Center, GNewSoft Co., Ltd.} \\
    }
  \thanks{*\,Corresponding author. Email: \texttt{ds.kim@gnewsoft.com}}
}

\begin{document}

\maketitle
\thispagestyle{empty}
\pagestyle{empty}

\begin{abstract}
Hyperspectral imagery provides rich spectral detail but poses unique challenges due to its high dimensionality in both spatial and spectral domains. \textbf{Therefore, we propose \emph{HyperspectralMAE}}, a transformer‑based foundation model for hyperspectral data featuring a \emph{dual‑masking} strategy that randomly occludes 50\% of the spatial patches and 50\% of the spectral bands during pre‑training. This forces the model to learn meaningful representations by reconstructing missing information across both dimensions. A positional embedding with spectral wavelength based on learnable harmonic Fourier components is introduced to encode the identity of each spectral band, ensuring that the model is sensitive to spectral order and spacing. The reconstruction objective employs a composite loss combining mean‑squared error (MSE) and spectral angle mapper (SAM) to balance pixel‑level accuracy and spectral‑shape fidelity.

The resulting model is of foundation scale ($\approx$ 0.18 B parameters, 768‑dimensional embeddings), indicating a high‑capacity architecture suitable for transfer learning. We evaluated \emph{HyperspectralMAE} on two large‑scale hyperspectral corpora—\textbf{NASA EO‑1 Hyperion} (1,600 scenes, about 300 billion pixel spectra) and \textbf{DLR EnMAP Level‑0} (1,300 scenes, about 300 billion pixel spectra)—and fine‑tuned it for land‑cover classification on the Indian Pines benchmark. Transfer‑learning results on Indian Pines demonstrate state‑of‑the‑art performance, confirming that our masked pre‑training yields robust spectral–spatial representations. The proposed approach highlights how dual masking and spectral embeddings can advance hyperspectral image reconstruction.
\end{abstract}

\section{INTRODUCTION}
\label{sec:introduction}

\IEEEPARstart{H}{yperspectral} images (HSIs) capture dozens to hundreds of contiguous spectral bands per pixel, enabling fine‑grained material discrimination in fields such as remote sensing, agriculture, and mineral exploration. However, this wealth of spectral information leads to high dimensionality and large data volume, making representation learning for HSIs challenging~\cite{zhu2023spectralmae,hong2022spectralformer}.

Self‑supervised foundation models—especially masked autoencoders (MAE)—have recently shown promise in learning robust representations from unlabeled data in computer vision~\cite{he2022mae}. In the hyperspectral domain, such models offer the potential to exploit rich spectral–spatial redundancies, yet existing work remains limited to small‑scale tasks. \textbf{Therefore, we propose \emph{HyperspectralMAE}}, a dedicated masked‑modeling framework that scales MAE principles to foundation‑model size for hyperspectral imagery.

\subsection{Motivation and Contributions}
Our main contributions are:
\begin{itemize}
    \item \textbf{Dual‑Masking Strategy:} We simultaneously mask spatial patches and spectral bands, forcing the model to capture cross‑dimensional dependencies~\cite{wang2024hsimae}.
    \item \textbf{Fourier‑Based Spectral Encoding:} Spectral wavelength embeddings are introduced via harmonic Fourier components, encouraging the model to learn spectral locality and periodicity~\cite{sohail2025energyformerenergyattentionfourier}.
    \item \textbf{Spectral–Spatial Reconstruction Loss:} A composite loss combining mean‑squared error (MSE) and spectral angle mapper (SAM)~\cite{kruse1993sam} ensures both numerical accuracy and high‑fidelity spectral profiles~\cite{Superpixel}.
    \item \textbf{Foundation‑Scale Model:} Our transformer architecture scales to hundreds of millions of parameters, setting a new size benchmark for hyperspectral foundation models.
\end{itemize}

\subsection{Organization}
The remainder of the paper is organized as follows. Section~\ref{sec:related} reviews related work on masked autoencoders and hyperspectral representation learning. Section~\ref{sec:method} details the proposed model and training methodology. Section~\ref{sec:experiments} describes the experimental setup and datasets. Section~\ref{sec:results} presents empirical results, and Section~\ref{sec:conclusion} concludes the paper.

\begin{figure*}[!t]         
  \centering
  \includegraphics[width=0.98\textwidth]{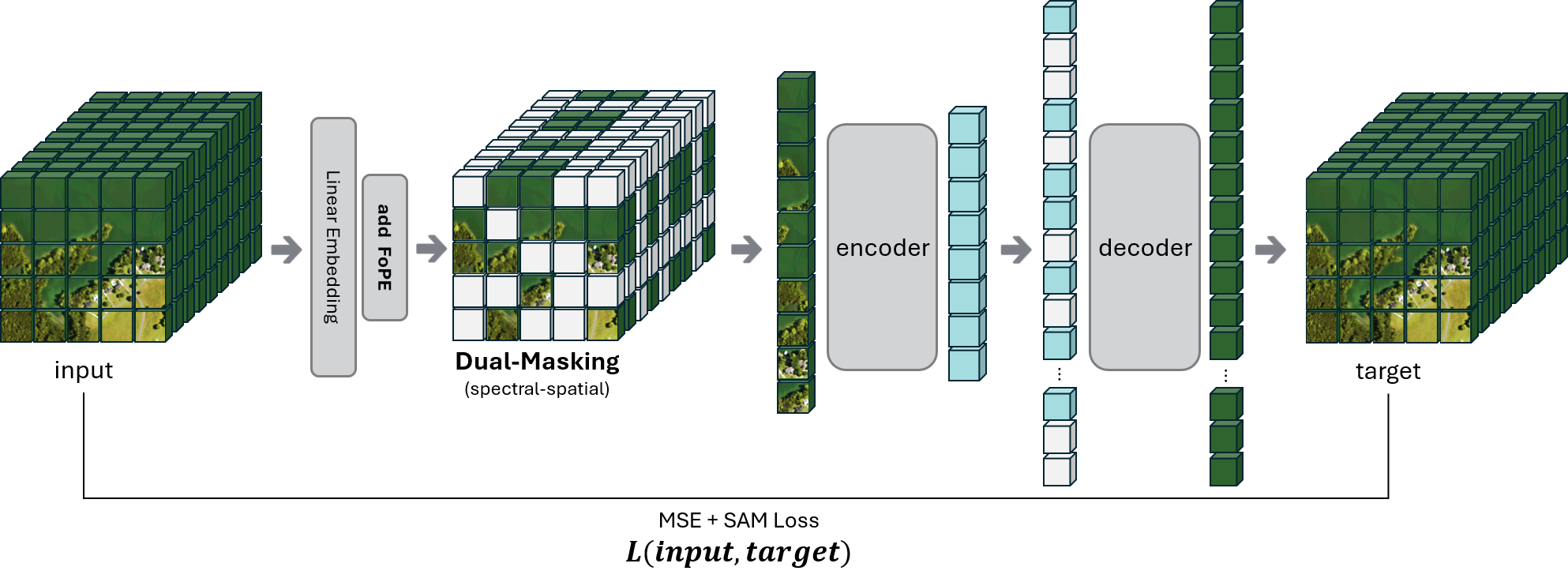}
  \caption{HyperspectralMAE overall pipeline.}
  \label{fig:overview}
\end{figure*}

\section{RELATED WORK}
\label{sec:related}

\subsection{Masked Autoencoders (MAE)}
Masked image modeling has gained popularity after the success of Vision MAE by He \emph{et al.}~\cite{he2022mae}, which showed that reconstructing masked patches is an effective pretext task for learning image representations. The original MAE applies random masking to input image patches and uses a ViT~\cite{dosovitskiy2021vit} encoder to encode visible patches and a lightweight decoder to reconstruct the missing patches. This approach has been extended to various domains.  

In hyperspectral imaging, a straightforward application is to mask spatial patches of the HSI cube and reconstruct them. Zhuang \emph{et al.}~\cite{Zhuang} applied an MAE‑like augmentation network for target detection in HSI, using a 1‑D positional encoding for spectra because 2‑D positional encoding was deemed unsuitable for spectral data. Similarly, Lin \emph{et al.}~\cite{lin2023ssmaespatialspectralmaskedautoencoder} introduced SS‑MAE, a spatial–spectral masked autoencoder with two branches: one masks and reconstructs random spatial patches, while the other masks and reconstructs random spectral bands. Their results confirm that incorporating spectral masking in addition to spatial masking leads to better feature learning. Our dual‑masking strategy is inspired by this idea, but we integrate both aspects within a single unified architecture and additionally introduce cross‑branch interactions.

\subsection{Transformers for HSI}
Vision Transformers (ViT)~\cite{dosovitskiy2021vit} have been applied to hyperspectral images in recent studies. A naïve approach treats each hyperspectral patch (with all bands) as a token sequence and applies ViT for classification. While ViTs can model long‑range spatial relationships, early attempts often ignored the explicit spectral structure. For example, some works reduced HSI to RGB (three bands) to apply pretrained ViTs, losing most spectral information. Others processed each pixel’s spectrum with a spectral transformer, treating each pixel’s spectral vector as a sequence of tokens.  

Scheibenreif \emph{et al.}~\cite{Scheibenreif} proposed a Spectral Transformer that attends to bands for each pixel, significantly improving accuracy by leveraging full spectral information. They further introduced a factorized Spatial–Spectral Transformer (SST), which applies self‑attention sequentially along spatial and spectral dimensions to handle the large token count in HSI. Factorizing attention in this way proved efficient and boosted performance. Our dual‑branch transformer shares a similar motivation of decoupling spectral and spatial attention, but instead of sequential factorization, we employ parallel branches with cross‑attention fusion to jointly learn spectral–spatial features.

\subsection{HSI‑Specific Networks}
Recently, various specialized networks have been designed explicitly for hyperspectral image (HSI) classification tasks. Traditionally, methods such as 3-D convolutional neural networks (3-D CNNs) and hybrid spectral-spatial CNNs have been widely adopted due to their ability to directly extract spatial-spectral features from HSI cubes via volumetric convolutions.

Emerging advancements have focused on incorporating transformer architectures into HSI-specific contexts to leverage self-attention mechanisms for improved representation learning. For example, Zhu \emph{et al.}~\cite{zhu2023spectralmae} developed SpectralMAE, a spectral-focused masked autoencoder explicitly designed for reconstructing spectral bands from partially masked inputs. SpectralMAE employs a specialized positional encoding tailored to spectral data, achieving superior reconstruction quality for various band combinations.

Similarly, Wang \emph{et al.}~\cite{wang2024hsimae} introduced HSI-MAE, which utilizes separate spatial and spectral encoders with fusion layers. HSI‑MAE explicitly learns spatial context and spectral correlation via two encoder branches, somewhat similar to our dual‑branch design. 

Our work, \emph{HyperspectralMAE} goes further by introducing learnable Fourier‑based positional features and a cross‑attention‑based embedding mechanism with MSE + SAM Loss function.  

In summary, whereas previous works have addressed spectral–spatial feature learning in parts (e.g., 1‑D positional embeddings or dual‑branch encoders), our work combines and extends these ideas in a unified framework with foundation model scale. We next describe the components of \emph{HyperspectralMAE} in detail.

\section{METHODOLOGY}
\label{sec:method}

We begin by outlining the encoder–decoder Transformer architecture of \emph{HyperspectralMAE}, which employs specialized embedding modules and dual masking operations. The encoder receives a partially masked hyperspectral image and produces latent representations, while the decoder reconstructs the full HSI from these latents. During pre‑training, a fraction of the input is masked both spatially and spectrally, forcing the encoder to infer missing information from the unmasked context along both dimensions. To this end, several key modules are introduced and detailed below.

\subsection{Overview of \emph{HyperspectralMAE} Architecture}
\figureautorefname{  \ref{fig:overview}} shows \emph{HyperspectralMAE}'s overall pipeline. Let some of $\mathbf{X}\in \mathbb{R}^{H \times W \times B}$, 
$$\mathbf{X}=\left\{x_{i, j, b} \in \mathbb{R} \mid i=1, \ldots, H ; j=1, \ldots, W ; b=1, \ldots, B\right\}$$
be an input hyperspectral image with spatial height $H$, width $W$, and $B$ spectral bands.
We partition X into non-overlapping spatial-spectral patches: each spatial patch covers a $9 \times 9$ pixel window and each spectral patch spans a block of 8 consecutive bands.
Formally, for spatial indices $p=1, \ldots,\lfloor H / 9\rfloor$ and $q=1, \ldots,\lfloor W / 9\rfloor$, and spectral-group index $k=1, \ldots,\lfloor B / 8\rfloor$, a patch is defined as
%%%
$$
\mathbf{P}_{p, q, k}
= \left\{ (i, j, b)
  \;\middle|\;
  \begin{aligned}
    9(p-1) &< i \le 9 p,\\
    9(q-1) &< j \le 9 q,\\
    8(k-1) &< b \le 8 k
  \end{aligned}
\right\}
$$
as HSI-MAE{~\cite{wang2024hsimae}} did.
This yields tokens that represent local spatial–spectral regions. Each token is augmented with learned positional encodings—2‑D spatial encodings (as in ViT) and our proposed spectral wavelength encodings for the band‑group index. The encoder, a ViT‑like Transformer, operates on visible unmasked tokens to produce latent features; a lightweight decoder then combines these latents with learned mask tokens to reconstruct the original hyperspectral image.

\subsection{Dual Spatial--Spectral Masking}

Inspired by HSI-MAE{~\cite{wang2024hsimae}, 
For spatial masking, we randomly sample $50 \%$ of spatial grid coordinates $(p, q)$ and mask all tokens whose spatial index equals any of the selected $(p, q)$ pairs-i.e., every spectral-group slice at those positions is hidden. and for spectral masking from the remaining tokens after spatial masking, we randomly sample $50 \%$ of spectral-group indices $k$ (equivalently, band groups $b$ ) and mask every token whose spectral index belongs to this set, regardless of its spatial location.

\subsection{Spectral Wavelength Positional Encoding}
In \emph{HyperspectralMAE}, we introduce a \emph{spectral wavelength} positional embedding to capture global spectral dependencies.
For each spectral patch $k$ (eight consecutive bands) we first compute its representative wavelength as the arithmetic mean of the centre-wavelengths of those bands:

$$
\lambda_k=\frac{1}{8} \sum_{m=1}^8 \lambda_{b_k^{(m)}},
$$
Can be corresponding angulaer frequency is
$$
\omega_k=\frac{2 \pi}{\lambda_k}.
$$

Where original{~\cite{vaswani2023attentionneed}} positional encoding(PE) is
$$
\begin{aligned}
\mathrm{PE}(\mathrm{pos}, 2 i) & =\sin \left(\frac{\mathrm{pos}}{10000^{2 i / d_{\text {model }}}}\right), \\
\mathrm{PE}(\mathrm{pos}, 2 i+1) & =\cos \left(\frac{\mathrm{pos}}{10000^{2 i / d_{\text {model }}}}\right),
\end{aligned}
$$
where $i$ indexes the dimension and $d_{model}$ is the embedding dimention.
We can analogously define a multi‐frequency spectral encoding that injects the wavelength $\lambda_k$ at multiple scales. Let $d_{spec}$ be our desired embedding dimension. For $i=0,1, \ldots, \frac{d_{\text {spec }}}{2}-1$, we define:

$$
\begin{aligned}
\operatorname{SpecEnc}\left(\lambda_k, 2 i\right) & =\sin \left(\frac{\omega_k}{10000^{\frac{2 i}{d_{\text {spec }}}}}\right), \\
\operatorname{SpecEnc}\left(\lambda_k, 2 i+1\right) & =\cos \left(\frac{\omega_k}{10000^{\frac{2 i}{d_{\text {dpec }}}}}\right),
\end{aligned}
$$
Hence, the spectral embedding vector for band b is
$$
\begin{aligned}
\mathbf{v}_{\text{spec}}(\lambda_k)
= \Bigl[
 &\sin\Bigl(\tfrac{\omega_k}{10000^0}\Bigr),
  \cos\Bigl(\tfrac{\omega_k}{10000^0}\Bigr), \\
 &\sin\Bigl(\tfrac{\omega_k}{10000^{2/d_{\text{spec}}}}\Bigr),
  \cos\Bigl(\tfrac{\omega_k}{10000^{2/d_{\text{spec}}}}\Bigr),
  \ldots
\Bigr].
\end{aligned}
$$
cycling over all $i$.

\subsection{Training Objective: MSE with SAM}

Let some of $\mathbf{Y}\in \mathbb{R}^{H \times W \times B}$,

$$
\mathbf{Y}=\left\{y_{i, j, b} \in \mathbb{R} \mid 1 \leq i \leq H, 1 \leq j \leq W, 1 \leq b \leq B\right\} 
$$

denote the ground-truth hyperspectral data cube, where $H$ and $W$ are the spatial dimensions and $B$ is the number of spectral bands.

The network produces a reconstruction

$$
\hat{\mathbf{Y}}=\left\{\hat{y}_{i, j, b}\right\} \in \mathbb{R}^{H \times W \times B}
$$

During pre-training, a two-stage masking procedure (Sect. C) selects a subset

$$
\mathcal{M} \subseteq\{1, \ldots, H\} \times\{1, \ldots, W\} \times\{1, \ldots, B\}
$$

of spatial-spectral indices to be hidden from the encoder.
For convenience, the complete set of spatial coordinates is written as

$$
\Omega=\{1, \ldots, H\} \times\{1, \ldots, W\}
$$
The first component of the loss is the mean squared reconstruction error evaluated only at masked voxels:

$$
\mathcal{L}_{\mathrm{MSE}}=\frac{1}{|\mathcal{M}|} \sum_{(i, j, b) \in \mathcal{M}}\left(y_{i, j, b}-\hat{y}_{i, j, b}\right)^2
$$

By confining the error computation to $\mathcal{M}$, the encoder-decoder is forced to infer radiance values that were never observed at input time.
3) Spectral-angle mapper over all pixels

For each pixel $(i, j) \in \Omega$, define the ground-truth and reconstructed spectra

$$
\mathbf{y}_{i, j}=\left[y_{i, j, 1}, \ldots, y_{i, j, B}\right]^{\top}, \quad \hat{\mathbf{y}}_{i, j}=\left[\hat{y}_{i, j, 1}, \ldots, \hat{y}_{i, j, B}\right]^{\top} .
$$

Their spectral angle is

$$
\operatorname{SAM}(i, j)=\arccos \left(\frac{\mathbf{y}_{i, j}^{\top} \hat{\mathbf{y}}_{i, j}}{\left\|\mathbf{y}_{i, j}\right\|_2\left\|\hat{\mathbf{y}}_{i, j}\right\|_2}\right)
$$

Averaging over all spatial locations produces the SAM loss

$$
\mathcal{L}_{\mathrm{SAM}}=\frac{1}{|\Omega|} \sum_{(i, j) \in \Omega} \operatorname{SAM}(i, j)
$$

This term promotes spectral-shape fidelity throughout the entire image rather than only at the masked positions, a property shown to benefit downstream analytical tasks.
The final training objective is a convex combination of the two criteria:

$$
\mathcal{L}_{\mathrm{rec}}=\alpha \mathcal{L}_{\mathrm{MSE}}+(1-\alpha) \mathcal{L}_{\mathrm{SAM}}
$$

where the weighting factor $\alpha \in[0,1]$ determines the trade-off between radiometric accuracy and spectral-shape preservation.
In all experiments we set $\alpha=0.5$.
Gradients originating from $\mathcal{L}_{\text {MSE }}$ are back-propagated exclusively through the masked voxels, whereas $\mathcal{L}_{\text {SAM }}$ distributes its gradients across the full data cube.
This asymmetric propagation compels the network to reconstruct unseen spatial-spectral regions with high numerical precision while maintaining globally coherent spectral signatures-yielding latent representations that are well-suited for subsequent hyperspectral classification task.

\section{EXPERIMENTS}
\label{sec:experiments}

\subsection{Pre-Training Data}
We assembled a large corpus of hyperspectral imagery to pre‑train \emph{HyperspectralMAE} in a self‑supervised manner. This corpus includes \textbf{NASA EO‑1 Hyperion}~\cite{folkman_eo1-hyperion_2000} and \textbf{DLR EnMAP Level‑0}~\cite{guanter_enmap_2015}, among others, totaling on the order of 600 billion raw pixels. These datasets span a variety of scenes and sensor characteristics, providing a diverse spectral–spatial training signal. We did not use any class labels during pre‑training; the model learned purely by reconstructing masked portions of the input data. We apply standard data augmentations such as random flipping and mild spectral jitter to increase robustness. The model was trained with an AdamW optimizer for several epochs until convergence of the reconstruction loss.

\bigskip
\noindent
\textbf{NASA EO‑1 Hyperion (1,600 scenes, $\sim$300 billion pixel spectra)}  
Each Hyperion scene covers a 7.7 km (cross‑track) $\times$ 42 km (along‑track) swath at 30 m spatial sampling ($\sim$256 $\times$ 1,400 pixels) over 220 contiguous spectral bands (0.357–2.576 $\mu$m, 10 nm sampling), totaling roughly 300 billion pixels.

\bigskip
\noindent
\textbf{DLR EnMAP Level‑0 (1,300 scenes, $\sim$300 billion pixel spectra)}  
Each EnMAP scene covers a 30 km $\times$ 30 km area at 30 m ground sampling (1,000 $\times$ 1,000 pixels) across 246 spectral bands (430–2,450 nm; 6.5–10 nm sampling), for a similar total of about 300 billion pixels.

\subsection{Model Scale and Implementation}
The \emph{HyperspectralMAE} encoder is a multi‑layer Transformer with 768‑dimensional embeddings and multi‑head self‑attention. A shallower decoder reconstructs the full HSI. The model totals about $1.8\times10^{8}$ parameters, enabled by large‑scale unlabeled data and mixed‑precision training.

\subsection{Fine-Tuning on Classification}
After pre‑training, we evaluate the learned representations on downstream HSI classification tasks. We evaluate on the \emph{Indian Pines} dataset—a classic hyperspectral scene of an agricultural area captured by the AVIRIS sensor, with 224 spectral bands and 16 ground‑truth land‑cover categories (after masking unused classes). Indian Pines is particularly challenging due to its limited size (only 10,249 labeled pixels) and high intra‑class spectral variability.

\subsection{Evaluation Metrics}
We evaluate on the standard Indian Pines dataset (145 $\times$ 145 pixels, 16 classes) following exactly the same train–test split and settings used by Hong \emph{et al.}~\cite{Hong2021SpectralFormer} for \emph{SpectralFormer}—a protocol subsequently adopted by HyLITE~\cite{Zhou2023HyLITE}—to enable direct comparison. All models are assessed with three common HSI metrics: Overall Accuracy (OA), Average Accuracy (AA), and Cohen’s Kappa coefficient ($\kappa$), which measure overall pixel correctness, mean per‑class accuracy, and chance‑corrected agreement with ground truth, respectively. By mirroring the prior evaluation protocol, we ensure that results for \emph{HyperspectralMAE} are directly comparable to existing methods under identical conditions.

\section{Results and Discussion}
\label{sec:results}

\subsection{Reconstruction Performance}
During self‑supervised pre‑training, \emph{HyperspectralMAE} demonstrated the ability to accurately reconstruct masked portions of the input. These indicate that the model learns a high‑fidelity internal representation of HSI data.

\subsection{Transfer Learning to Indian Pines}

The Indian Pines results highlight the effectiveness of \emph{HyperspectralMAE} as a foundation model for HSI. With only small training pixels the fine‑tuned model achieved an Overall Accuracy (OA) of \textbf{92.37\%}, surpassing previous state‑of‑the‑art methods. Compared with the best conventional CNN baseline (2‑D CNN \cite{Chen2016}, 75.89\% OA) and the transformer‑based SpectralFormer ~\cite{Hong2021SpectralFormer} (78.97\% OA), \emph{HyperspectralMAE} shows gains of roughly 13 percentage points in OA and 10 points in AA, establishing a new benchmark for this dataset. The result shown at \tableautorefname{  \ref{tab:indian_pines_results}}.

These substantial gains emphasize the strength of our pre‑training and fine‑tuning strategy, particularly when training data are limited.

\begin{table}[!htbp]
  \renewcommand{\arraystretch}{1.2}
  \caption{Classification Results on the Indian Pines Dataset}
  \centering
  \begin{tabular}{lccc}
    \hline
    \textbf{Method} & \textbf{OA (\%)} & \textbf{AA (\%)} & \textbf{Kappa} \\
    \hline
    kNN~\cite{Cover1967}           & 59.17 & 63.90 & 0.54 \\
    RF~\cite{Breiman2001}          & 69.80 & 76.78 & 0.65 \\
    SVM~\cite{Cortes1995}          & 72.36 & 83.16 & 0.68 \\
    1‑D CNN~\cite{Chen2016}        & 70.43 & 79.60 & 0.66 \\
    2‑D CNN~\cite{Chen2016}        & 75.89 & 86.64 & 0.72 \\
    RNN~\cite{Hang2019}            & 70.66 & 76.37 & 0.66 \\
    miniGCN~\cite{Hong2021GCN}     & 75.11 & 78.03 & 0.71 \\
    ViT~\cite{dosovitskiy2021vit}  & 71.86 & 78.97 & 0.68 \\
    SpectralFormer~\cite{Hong2021SpectralFormer} & 78.97 & 85.39 & 0.76 \\
    HyLITE~\cite{Zhou2023HyLITE}   & 89.80 & 94.69 & 0.88 \\
    \hline
    \textbf{Ours (\emph{HyperspectralMAE})} & \textbf{92.37} & \textbf{95.80} & \textbf{0.91} \\
    \hline
  \end{tabular}
  \label{tab:indian_pines_results}
\end{table}

\section{CONCLUSIONS}
\label{sec:conclusion}

We introduced \emph{HyperspectralMAE}, a transformer‑based masked autoencoder that learns rich spectral–spatial representations from hyperspectral imagery. The proposed \emph{dual‑masking} strategy—simultaneously hiding 50\,\% of spatial patches and 50\,\% of spectral bands—forces the network to reason jointly across both dimensions. A wavelength‑aware spectral positional embedding further equips the model with an explicit notion of spectral frequency, while a composite MSE\,+\,SAM reconstruction loss balances pixel‑level accuracy and spectral‑shape fidelity. Scaled to about 0.18\,B parameters, \emph{HyperspectralMAE} attains state‑of‑the‑art results on Indian Pines, even when fine‑tuned with only few labeled pixels.

These findings underscore the promise of large‑scale self‑supervised learning for hyperspectral data, a domain often constrained by scarce annotations. The model’s strong transfer performance on limited supervision suggests practical value for real‑world applications such as precision agriculture, environmental monitoring, and mineral exploration.

Future work will explore scaling \emph{HyperspectralMAE} to even larger model sizes and extending the framework to additional tasks, including anomaly detection, spectral unmixing, and retrieval. We believe this study represents a step toward general‑purpose foundation models for hyperspectral sensing, bridging the gap between rich spectral data and modern representation learning.
\bibliographystyle{IEEEtran}
\bibliography{
references}

\end{document}